\documentclass[conference]{IEEEtran}
\IEEEoverridecommandlockouts
\usepackage{cite}
\usepackage{amsmath,amssymb,amsfonts}
\usepackage{algorithmic}
\usepackage{graphicx}
\usepackage{textcomp}
\usepackage{booktabs}
\usepackage{xcolor}
\usepackage{url}
\usepackage{amsmath}
\usepackage{booktabs}
\usepackage{pifont} 

\usepackage[linesnumbered,ruled,vlined]{algorithm2e}
\usepackage{setspace}
\SetKwComment{tcp}{$\triangleright\ $}{} 
\usepackage{multirow}
\usepackage{pgfplots} 
\usepackage{mwe}
\usepackage{hyperref}
\def\BibTeX{{\rm B\kern-.05em{\sc i\kern-.025em b}\kern-.08em
    T\kern-.1667em\lower.7ex\hbox{E}\kern-.125emX}}
\begin{document}

\title{Physics-Encoded Inverse Modeling for Arctic Snow Depth Estimation\\
}
\author{
    \IEEEauthorblockN{Akila Sampath\IEEEauthorrefmark{1}\IEEEauthorrefmark{2}, Vandana P. Janeja\IEEEauthorrefmark{1}\IEEEauthorrefmark{2}, and Jianwu Wang\IEEEauthorrefmark{1}\IEEEauthorrefmark{2}}
    \IEEEauthorblockA{\IEEEauthorrefmark{1}Department of Information Systems, University of Maryland Baltimore County,}
    \IEEEauthorblockA{Baltimore, MD, USA}
    \IEEEauthorblockA{\IEEEauthorrefmark{2}iHARP: NSF HDR Institute for Harnessing Data and Model Revolution in the Polar Regions}
    \IEEEauthorblockA{\{asampath, jianwu, vjaneja\}@umbc.edu}
}
\maketitle

\begin{abstract}

Accurate estimation of unobserved quantities in time-varying inverse problems remains challenging when observations are sparse and only indirectly related to the target variable. In Arctic climate applications, snow depth over sea ice is not directly available in commonly used reanalysis products and must instead be inferred from related physical and environmental variables. To address this challenge, we introduce Physics-Encoded Inverse Modeling (PhysE-Inv), a framework that combines sequential deep learning with a physics-encoded parameter estimation module for inverse estimation under sparse observational conditions. PhysE-Inv uses an LSTM encoder-decoder to capture temporal dependencies and incorporates contrastive learning to improve the consistency of learned representations. The framework learns structured physics-encoded parameters that are integrated with observational inputs to estimate snow depth proxies. Under the proposed proxy evaluation framework, PhysE-Inv outperforms all evaluated baselines, achieving an average MSE reduction of 24.4\% compared with baseline models and a 17.3\% improvement over the strongest baseline under the parameter estimation setting. These results demonstrate the potential of physics-encoded modeling approaches for estimating unobserved quantities in data-scarce domains where direct observations are limited.
\end{abstract}
\section{Introduction}
The Arctic Ocean presents a complex, partially observable system where key physical quantities remain unobserved. For example, Arctic snow depth is unobserved in ERA5 reanalysis \cite{Hersbach}. We aim to demonstrate how sparse, indirectly related observations can be used to estimate a physically motivated snow depth proxy, approached from an inverse-problem perspective. Inverse modeling has proven valuable for inferring hidden physical parameters across diverse scientific applications \cite{Tayal2022,sun2020theory,ghorbanidehno2020recent}. However, existing complex inverse-forward transformations \cite{tarantola2005inverse,Ghosh2022Robust,Tayal2022} typically require a computationally expensive perfect invertibility assumption between the input and latent spaces. A one-to-one correspondence works well for static systems, but fails to capture time-varying physical systems where underlying parameters evolve dynamically.

Not all physical systems require complex partial differential equations (PDEs) and Physics-Informed Neural Networks (PINNs) \cite{faroughi2022physics} to embed physical structure. Systems governed by linear structure, such as the hydrostatic equilibrium relationship between sea ice and snow \cite{kwok2008icesat}, can be effectively modeled using simpler physics-encoded approaches. Beyond inverse modeling, self-supervised learning (SSL) has shown promise for representation learning in data-scarce scientific settings \cite{scotti2023reconstructing, liu2021self, jing2020self}, though current SSL approaches often rely on generative or regularization losses requiring extensive tuning, or assume bijective mappings (perfect invertibility assumption) \cite{Ghosh2022Robust}, limiting applicability to noisy, incomplete real-world observations.

The main contributions of this work are as follows. We introduce Physics-Encoded Inverse Modeling (PhysE-Inv), a unified framework that integrates sequential modeling, self-supervised contrastive regularization, and physics-encoded parameter estimation for data-scarce Arctic snow depth estimation. Contrastive regularization is incorporated directly into training, rather than as a standalone pretraining step. The framework learns bounded, structured parameters through differentiable transformations. These parameters are combined with data-driven predictions through a reconstruction equation motivated by, but not derived from, hydrostatic equilibrium. We demonstrate the effectiveness of PhysE-Inv through extensive benchmarking, showing improved predictive performance and robustness under sparse data conditions.
\section{Related Works}

Physics-Informed Neural Networks (PINNs) represent a robust paradigm for integrating scientific principles into machine learning frameworks \cite{faroughi2022physics, Willard2022, karpatne2024knowledge}. By incorporating physical laws through differential equations and other mathematical constraints \cite{rao2021hard, kovachki2021neural, innes2019differentiable}, these models aim to improve predictive performance while adhering to physical principles, moving beyond purely black-box models. Previous studies \cite{raissi2019physics,nguyen2025physixfoundationmodelphysics, Lu_2021} have been specifically tailored for solving or learning complex nonlinear partial differential equations (PDEs) or ordinary differential equations (ODEs). While powerful for numerical simulation and system identification, this machinery is often ill-suited for the simpler, linear relationships considered in this work. Where the goal is to infer an unobserved quantity from incomplete observations, PDE-based PIML approaches are often computationally expensive and less suited to sparse-data settings. Similarly, foundation models such as ClimaX \cite{nguyen2023climaxfoundationmodelweather} are designed for large-scale climate prediction and cross-variable representation learning in data-rich environments, rather than the data-scarce, single-target estimation problems considered in this work. 

Inverse modeling has emerged as a powerful framework for inferring unobserved physical quantities in complex geospatial systems from sparse and indirect observations. This approach has gained traction across domains including hydrology \cite{Ghosh2022Robust}, water flow modeling \cite{mo2020deep}, and lake temperature estimation \cite{Tayal2022}. In seismic waveform inversion, wave propagation theory has been integrated into deep learning frameworks \cite{Adler2021Deep}, and deep neural networks have been applied to electrical impedance tomography (EIT) by approximating the inverse of the nonlinear, high-dimensional Dirichlet-to-Neumann (DtN) map \cite{fan2020solving}. Similarly, architectures like SwitchNet handle both forward and inverse wave equation scattering, capturing global phenomena with high computational efficiency \cite{khoo2021switchnet}.

Our work differs from these approaches in scope and design choices rather than in formal invertibility guarantees. Rather than assuming a bijective, perfectly invertible mapping between observations and the target, we adopt a physics-encoded parameter estimation module combined with contrastive learning. We do not formally instantiate a forward operator or prove invertibility properties. Instead, we treat this as motivation for a learned structural prior and evaluate its practical effectiveness under real-world Arctic data scarcity.

\section{Methodology}
\subsection{Problem Formulation and Preliminaries}

The primary input is a time series of snow density measurements $\mathbf{X} = [x_1, x_2, \dots, x_n]^\top \in \mathbb{R}^{n \times 1}$, where $x_t$ denotes the normalized snow density at time step $t$ and $n$ is the sequence length ($n=10$ in our experiments). We generate an augmented sequence $\mathbf{X}' = [x'_1, x'_2, \dots, x'_n]^\top$ by perturbing the input with Gaussian noise:
\begin{equation}
x'_t = x_t + \epsilon_t,
\qquad
\epsilon_t \sim \mathcal{N}(0,\sigma^2),
\qquad
\forall t \in \{1,\dots,n\},
\end{equation}
\noindent
where $\sigma = 0.01$. The target variable is a normalized snow depth proxy sequence $\mathbf{Y} = [y_1, y_2, \dots, y_n]^\top \in \mathbb{R}^{n \times 1}$, derived from ERA5 variables via a physically motivated proxy, as described below. This proxy is used because direct snow depth observations are unavailable in ERA5 \cite{Hersbach}.

\subsection{Proxy Target Generation}
We formulate snow depth estimation as an inverse problem, where the objective is to infer the unobserved snow depth $h_{\mathrm{s}}$ from available observations of snow density and sea ice properties. True hydrostatic balance for a snow-covered ice floe is given by
\begin{equation}\label{hydro_balance}
\rho_{\mathrm{i}} h_{\mathrm{i}} + \rho_{\mathrm{s}} h_{\mathrm{s}}
= \rho_{\mathrm{w}} \left( h_{\mathrm{i}} - h_{\mathrm{f}} \right),
\end{equation}
\noindent
where $\rho_{\mathrm{i}}$, $\rho_{\mathrm{s}}$, and $\rho_{\mathrm{w}}$ denote the densities of ice, snow, and seawater, respectively, and $h_{\mathrm{i}}$, $h_{\mathrm{s}}$, and $h_{\mathrm{f}}$ represent ice thickness, snow depth, and freeboard. Solving Eq.~\ref{hydro_balance} for $h_{\mathrm{s}}$ requires ice thickness and freeboard as inputs. Since these variables are not available in ERA5, the target cannot be directly derived from Eq.~\ref{hydro_balance} using the available data.

We therefore construct $h_{\mathrm{s}}^{\mathrm{proxy}}$ as a heuristic, dimensionless proxy, motivated by but not derived from Eq.~\ref{hydro_balance}, using the ERA5 variables available to us: sea ice concentration ($C$), snow albedo ($\alpha_{\mathrm{s}}$), and snow density ($\rho_{\mathrm{s}}$):
\begin{equation}\label{eq:proxy_model}
h_{\mathrm{s}}^{\mathrm{proxy}}
=
\frac{C \cdot A + \alpha_{\mathrm{s}} \, \rho_{\mathrm{s}}}
{A-B},
\end{equation}
\noindent
where $A=600$ and $B=300$ are empirically selected scaling constants chosen to maintain comparable scales among the proxy components. The resulting $h_{\mathrm{s}}^{\mathrm{proxy}}$ is treated as a dimensionless proxy signal that captures relative snow variability rather than absolute snow depth. A more rigorous calibration of the proxy formulation is left for future work.

The learning objective is to approximate the inverse mapping
\begin{equation}
h_{\mathrm{s}}^{\mathrm{proxy}} \approx \mathcal{F}^{-1}(\rho_{\mathrm{s}}, \boldsymbol{\theta}),
\end{equation}
\noindent
where $\mathcal{F}^{-1}$ is parameterized by neural network weights $\boldsymbol{\theta}$, trained against $h_{\mathrm{s}}^{\mathrm{proxy}}$ as a physically-motivated but heuristic target. As $h_{\mathrm{s}}^{\mathrm{proxy}}$ correlates with, but does not numerically equal, true snow depth, $\mathcal{F}^{-1}$ should be understood as approximating this proxy quantity rather than $h_{\mathrm{s}}$ directly.

\subsection{Model Architecture}

\begin{figure*}[t] 
\centering 
\includegraphics[width=0.70\linewidth]{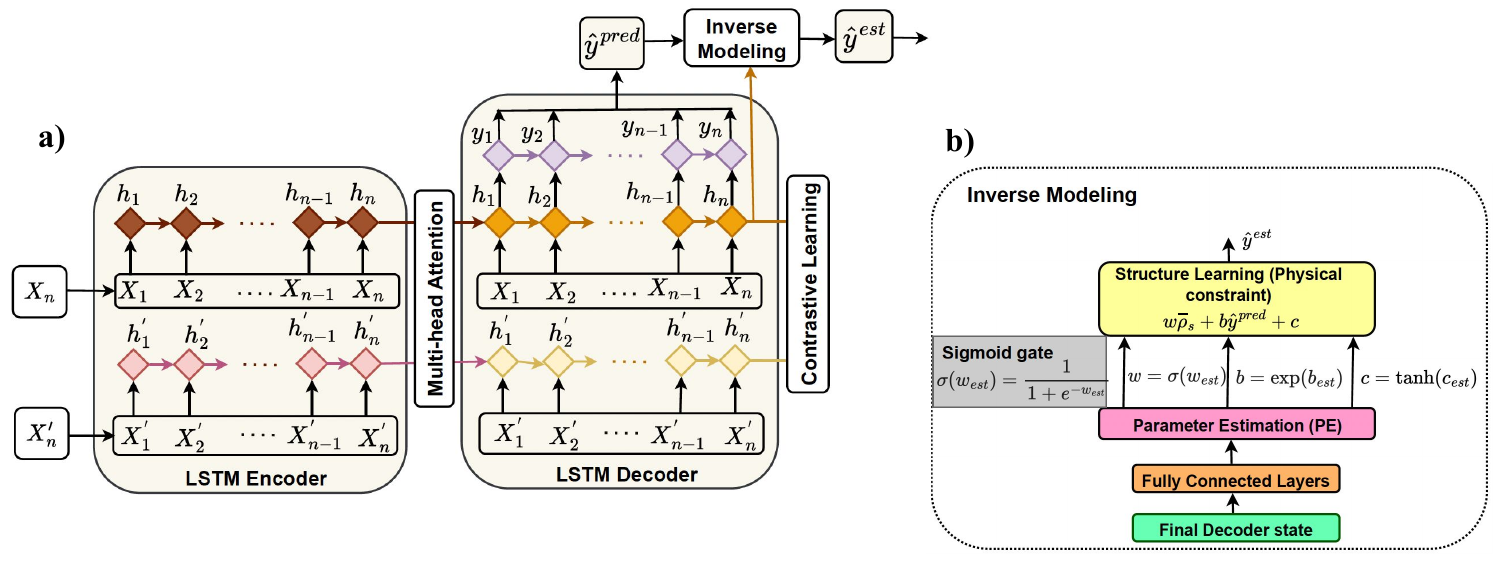} 
\caption{The Physics-Encoded Inverse Modeling (PhysE-Inv) framework integrates two key components: (a) Contrastive Learning for noise-robust representations, and (b) an inverse modeling framework that employs Parameter Estimation (PE).}
\label{fig:architecture}
\end{figure*}

The Physics-Encoded Inverse Modeling (PhysE-Inv) framework is illustrated in Figure~\ref{fig:architecture}. It integrates inverse sequence modeling, attention-based temporal feature refinement, physics-encoded parameter estimation, and contrastive regularization within a unified, end-to-end differentiable architecture.

\subsubsection{Encoder-Decoder Architecture}

\paragraph{LSTM Encoder}
Temporal dependencies within each sequence are captured using a two-layer Long Short-Term Memory (LSTM) encoder with hidden dimension 64 and dropout rate 0.4. At each time step $t \in \{1, \dots, n\}$, the encoder updates its hidden and cell states according to
\begin{equation}
\mathbf{h}_t^{\text{enc}}, \mathbf{c}_t^{\text{enc}} =
\text{LSTM}^{\text{enc}}(x_t, \mathbf{h}_{t-1}^{\text{enc}}, \mathbf{c}_{t-1}^{\text{enc}}),
\end{equation}
resulting in a sequence of hidden states
\begin{equation}
\mathbf{H}^{\text{enc}} = [\mathbf{h}_1^{\text{enc}}, \mathbf{h}_2^{\text{enc}}, \dots, \mathbf{h}_n^{\text{enc}}],
\end{equation}
This encoder provides a compact temporal embedding that summarizes both short-term fluctuations and longer-term trends in the input signal, capturing the temporal evolution of snow density across the seasonal cycle.

\paragraph{Multi-head Self-Attention for Temporal Refinement}
To capture non-local temporal interactions under sparse observations, encoder outputs $\mathbf{H}^{\text{enc}}$ are refined via a 4-head self-attention mechanism. For each head $i \in \{1, \dots, 4\}$, queries, keys, and values are projected as $\mathbf{Q}_i = \mathbf{H}^{\text{enc}}\mathbf{W}_i^Q$, $\mathbf{K}_i = \mathbf{H}^{\text{enc}}\mathbf{W}_i^K$, and $\mathbf{V}_i = \mathbf{H}^{\text{enc}}\mathbf{W}_i^V$, yielding:
\begin{equation}
\text{head}_i = \text{softmax}\left(\frac{\mathbf{Q}_i \mathbf{K}_i^\top}{\sqrt{\mathstrut d_k}}\right)\mathbf{V}_i
\end{equation}
\noindent
where $d_k$ denotes the key dimensionality, shared across all attention heads.
\begin{equation}
\begin{split}
\mathbf{A} &= \text{Concat}(\text{head}_1, \dots, \text{head}_4)\mathbf{W}^O, \\
\mathbf{H}^{\text{attn}} &= \text{LayerNorm}\big(\mathbf{H}^{\text{enc}} + \text{Dropout}(\mathbf{A})\big).
\end{split}
\label{eq:attention_refinement}
\end{equation}
This residual framework ensures stable downstream predictions and robust gradient flow across long-range dependencies.

\paragraph{LSTM Decoder}
The attention-enhanced representations are passed to a two-layer LSTM decoder,
\begin{equation}
\mathbf{h}_t^{\text{dec}}, \mathbf{c}_t^{\text{dec}} =
\text{LSTM}^{\text{dec}}(\mathbf{h}_t^{\text{attn}}, \mathbf{h}_{t-1}^{\text{dec}}, \mathbf{c}_{t-1}^{\text{dec}}),
\end{equation}
\noindent
which integrates refined temporal information into a coherent latent representation. The decoder processes the attention-enhanced sequence to produce hidden states at each time step, with the final decoder hidden state,
\begin{equation}
\mathbf{z}_n = \mathbf{h}_n^{\text{dec}},
\end{equation}
serving as a global summary of the entire input sequence. This final decoder state forms the basis for both direct prediction and inverse modeling via PE, acting as the compressed representation from which the heuristic parameters $(w,b,c)$ are inferred through a differentiable parameter estimation network.
\paragraph{Direct Prediction Head}
A linear projection applied to each decoder hidden state produces intermediate depth predictions,
\begin{equation}
\hat{y}_t^{\text{pred}} = \mathbf{W}_{\text{depth}}\mathbf{h}_t^{\text{dec}} + b_{\text{depth}}, \quad t \in \{1, \dots, n\},
\end{equation}
This direct prediction serves as a data-driven baseline that captures patterns learned from the training distribution.

\subsubsection{Parameter Estimation and Structure Learning}
\label{sec:physics_encoding}
Snow depth is not directly observed in Arctic reanalysis datasets such as ERA5, and the available variables, such as sea ice concentration and snow density, are only indirectly related to it. We therefore frame snow depth estimation from an inverse-problem perspective, though we do not instantiate an explicit forward operator or enforce invertibility as a formal constraint. Instead, we introduce a physics-encoded parameter module $(w, b, c)$ that provides a soft, learned structural prior, intended to discourage the model from relying on spurious data.
\paragraph{Parameter Estimation Module}
A 3-layer fully connected network maps the final latent state to raw parameters, $\mathbf{p}_{\text{raw}} = \text{FC}(\mathbf{z}_n) \in \mathbb{R}^3$. The bounded parameter vector $[w, b, c]^\top$ is then obtained via differentiable transformations:
\begin{equation}
[w, b, c] = \big[2\sigma(w_{\text{raw}}) - 1, \,\, \exp(b_{\text{raw}}), \,\, 10\tanh(c_{\text{raw}})\big],
\label{eq:pe_constraints}
\end{equation}
\noindent
where $\sigma(\cdot)$ is the sigmoid function. These transformations constrain the parameter ranges while maintaining numerical stability and preserving end-to-end gradient flow.

The parameters are estimated dynamically for each input sequence and admit physically motivated, though not independently verified, interpretations. Specifically, $w \in [-1,1]$ can be loosely interpreted as a density-depth coupling coefficient, with its sign suggesting compaction ($w>0$) or depth hoar formation ($w<0$). The parameter $b \in (0,\infty)$ acts as a confidence scaling factor that modulates the data-driven prediction, while $c \in [-10,10]$ provides a systematic bias correction for unmodeled atmospheric or seasonal processes.

These interpretations are intended to motivate rather than enforce physical constraints. Since $b$ and $c$ together already allow an affine transformation of $\hat{y}_t^{\mathrm{pred}}$, and $w$ is free to take either sign, the parameter estimation module functions as a learned structural prior rather than a hard physical constraint on the solution space.
\paragraph{Structure Learning}
The estimated parameters are integrated through a physics-inspired reconstruction equation:
\begin{equation}
\hat{y}^{\text{est}}_t = w\,\bar{\rho}_s + b\,\hat{y}^{\text{pred}}_t + c, \quad t = 1, \dots, T,
\label{eq:structure_learning}
\end{equation}
\noindent
where $\bar{\rho}_s = \frac{1}{n}\sum_{t=1}^{n} x_t$ denotes the mean input snow density. This formulation is motivated by, but not directly derived from, the hydrostatic balance equation (Eq.~\ref{hydro_balance}). It combines a density-based prior term ($w\bar{\rho}_s$), a confidence-scaled data-driven prediction ($b\hat{y}^{\text{pred}}_t$), and a systematic bias correction ($c$).

The relationship between snow depth and snow density is not governed by a fixed physical law and can vary under processes such as compaction, wind packing, and depth hoar formation. Consequently, the term $w\bar{\rho}_s$ should be interpreted as a learned linear feature with a physically motivated interpretation rather than an encoded physical law, with $w$ learned freely from the data. We note further that $b$ and $c$ together already allow this equation to apply a full affine transformation to $\hat{y}_t^{\text{pred}}$; the physics-inspired structure therefore functions as a learned prior rather than a hard constraint on the solution space.

The learned affine transformation also maps the dimensionless proxy signal $h_{\mathrm{s}}^{\mathrm{proxy}}$ (Eq.~\ref{eq:proxy_model}) to snow depth estimates through end-to-end optimization rather than an analytically prescribed conversion. By retaining a linear reconstruction model, the estimated parameters remain structurally interpretable while avoiding unnecessary nonlinear complexity.

\subsubsection{Contrastive Learning}
\label{sec:contrastive}
To stabilize the latent manifold under observational uncertainty, we employ an NT-Xent contrastive loss following \cite{chen2020simple}, which encourages clean and mildly perturbed versions of the same sequence ($\sigma = 0.01$, an approximately 1\% perturbation) to map to nearby points in the latent space, promoting local representation stability rather than demonstrated robustness to larger, more realistic noise levels. For a batch size $M$, we extract the final decoder states for the original input $\mathbf{z}_i = \mathbf{h}_n^{\text{dec}}(\mathbf{X}_i)$ and its perturbed counterpart $\mathbf{z}'_i = \mathbf{h}_n^{\text{dec}}(\mathbf{X}'_i)$. Following $\ell_2$-normalization, $\tilde{\mathbf{z}}_i = \mathbf{z}_i/\|\mathbf{z}_i\|_2$ and $\tilde{\mathbf{z}}'_i = \mathbf{z}'_i/\|\mathbf{z}'_i\|_2$, we form the joint representation matrix $\tilde{\mathbf{Z}} = [\tilde{\mathbf{z}}_1, \dots, \tilde{\mathbf{z}}_M, \tilde{\mathbf{z}}'_1, \dots, \tilde{\mathbf{z}}'_M] \in \mathbb{R}^{d \times 2M}$. The contrastive loss is defined as:
\begin{equation}
\mathcal{L}_{\text{contrast}} = -\frac{1}{2M} \sum_{i=1}^{2M} \log \frac{\exp(\tilde{\mathbf{z}}_i^\top \tilde{\mathbf{z}}_{\text{pair}(i)} / \tau)}{\sum_{j \neq i} \exp(\tilde{\mathbf{z}}_i^\top \tilde{\mathbf{z}}_j / \tau)},
\label{eq:contrastive_loss}
\end{equation}
\noindent
where $\text{pair}(i) = i + M$ if $i \leq M$ else $i - M$, and $\tau = 0.05$ is the temperature hyperparameter.

\subsection{Integrated Loss Formulation}
The complete training objective combines three loss terms:
\begin{equation}
\mathcal{L}_{\text{total}} = \mathcal{L}_{\text{pred}} + \mathcal{L}_{\text{physics}} + \lambda_{\text{contrast}}\mathcal{L}_{\text{contrast}},
\label{eq:total_loss}
\end{equation}
\noindent
where $\mathcal{L}_{\text{pred}} = \frac{1}{n}\sum_{t=1}^{n}(\hat{y}_t^{\text{pred}} - y_t^{\text{proxy}})^2$ supervises direct neural predictions, $\mathcal{L}_{\text{physics}} = \frac{1}{n}\sum_{t=1}^{n}(\hat{y}_t^{\text{est}} - y_t^{\text{proxy}})^2$ enforces consistency with Eq.~\ref{eq:structure_learning}, and $\mathcal{L}_{\text{contrast}}$ (Eq.~\ref{eq:contrastive_loss}) regularizes the latent representation weighted by the hyperparameter $\lambda_{\text{contrast}}$. Here, $y_t^{\text{proxy}}$ denotes the per-timestep value of the heuristic proxy target $h_{\mathrm{s}}^{\mathrm{proxy}}$ defined in Eq.~\ref{eq:proxy_model}.

\section{Experiments}
\subsection{Experimental Setup}
\subsubsection{Dataset}
This study uses the ERA5 atmospheric reanalysis dataset from the European Centre for Medium-Range Weather Forecasts (ECMWF) \cite{Hersbach}. ERA5 combines observational data, primarily from satellite remote sensing, to provide consistent daily climate fields. We extracted data over the central Arctic Ocean (70$^{\circ}$N–85$^{\circ}$N) from January 2019 to December 2023, yielding 1,826 daily time steps at a spatial resolution of 0.25$^{\circ}$ $\times$ 0.25$^{\circ}$ (approximately 25 km). Snow albedo, snow density, and sea ice concentration variables were used. Daily time series were generated by spatially averaging grid-point values and then normalized using statistics computed from the training set. The implementation is available \href{https://github.com/akilasampath5/PhysE-Inv}{here}.
\subsubsection{Implementation Details}
PhysE-Inv employs a two-layer LSTM encoder-decoder with 64 hidden units per layer and four-head self-attention. The model is trained for 500 epochs using Adam with a learning rate of $5\times10^{-4}$ and a batch size of 16. Dropout (0.4), L1 regularization ($\lambda=0.001$), and L2 weight decay ($1\times10^{-5}$) are used to reduce overfitting. The multi-task objective combines MSE losses for direct and physics-encoded estimates with a contrastive loss ($\tau=0.05$). Unless otherwise stated, all experiments use a fixed random seed of 42.
\subsubsection{Baselines}
To isolate the specific contributions of PhysE-Inv, our experimental design compares four temporal architectures: a standard recurrent LSTM baseline to isolate the value of explicit physics constraints, a BiLSTM to evaluate the impact of bidirectional context, a continuous-time Neural ODE to compare implicit differential equation learning with explicit physics-encoded parameter estimation, and a 1D ResNet-50 to compare hierarchical convolutional feature extraction with recurrent temporal modeling. For controlled comparisons, the PE-enhanced baseline models use the same physics-encoded parameter head and contrastive regularization strategy, while retaining their original backbone architectures.
\subsection{Results}
\subsubsection{Parameter Estimation (PE) Results}
PhysE-Inv is compared against four representative temporal modeling architectures under two configurations: (1) standard data-driven estimation without parameter estimation (PE), and (2) the same architectures augmented with the proposed physics-encoded parameter estimation module defined in Eq.~\ref{eq:structure_learning}. 

\begin{table}[t]
\centering
\caption{Model comparison across baseline configurations ($\sim$90\%/10\% train/test split). The last column shows \% MSE reduction of PhysE-Inv relative to each baseline (With PE).}
\label{tab:model_comparison}
\small
\setlength{\tabcolsep}{4pt}
\begin{tabular}{l c c c c c c} 
    \toprule
    \textbf{Model}
    & \multicolumn{2}{c}{\textbf{Without PE}}
    & \multicolumn{2}{c}{\textbf{With PE}}
    & \multicolumn{2}{c}{\textbf{MSE $\downarrow$ (\%)}} \\ 
    \cmidrule(lr){2-3} \cmidrule(lr){4-5} \cmidrule(lr){6-7}
    & \textbf{MSE} & \textbf{RMSE}
    & \textbf{MSE} & \textbf{RMSE}
    & \textbf{Reduction} \\
    \midrule
    LSTM         & 0.4679 & 0.6840 & 0.4545 & 0.6742 & 21.5 \\
    NeuralODE    & 0.5066 & 0.7117 & 0.4926 & 0.7018 & 27.6 \\
    ResNet50     & 0.4308 & 0.6563 & 0.4315 & 0.6569 & 17.3 \\
    BiLSTM       & 0.5263 & 0.7255 & 0.5177 & 0.7195 & 31.1 \\
    \textbf{PhysE-Inv}
                 & \textbf{0.3942} & \textbf{0.6278}
                 & \textbf{0.3568} & \textbf{0.5973} 
                    & -- \\
    \bottomrule
\end{tabular}
\end{table}
This analysis suggests that the effectiveness of physics-encoded parameter estimation depends on the underlying architecture. LSTM and NeuralODE exhibit notable reductions in MSE (2.9\% and 2.8\%, respectively), while BiLSTM shows more modest gains (1.6\%). In contrast, ResNet50 exhibits a slight increase in MSE (0.4308 to 0.4315), suggesting that the benefit of the proposed physics encoding may depend on how effectively the backbone architecture captures temporal dependencies.

PhysE-Inv achieves the best overall performance under both evaluation settings. Without parameter estimation, PhysE-Inv outperforms the strongest baseline (ResNet50) by 8.5\% in MSE, indicating that the proposed encoder-decoder architecture with multi-head attention provides effective temporal representation learning for sparse inverse inference tasks. When augmented with parameter estimation, PhysE-Inv further reduces MSE from 0.3942 to 0.3568 (9.5\%) and outperforms all baseline models under the PE-enabled setting by an average of 24.4\% in MSE.

\subsubsection{Ablation study: Impact of contrastive learning}

Sparse Regime I, II, and III correspond to training with 80\%, 60\%, and 50\% of the available data, respectively, using a chronological train/test split in which sliding windows are constructed independently within each partition, ensuring no temporal overlap between training and test samples. Note that the test set size also varies across regimes (20\%, 40\%, and 50\%, respectively). Table~\ref{tab:ablation_ratios} evaluates the contribution of contrastive learning under these varying sparse data regimes. 

\begin{table}[t]
    \centering
    \caption{Ablation study of PhysE-Inv performance under varying degrees of data sparsity with and without contrastive learning (CL).}
    \label{tab:ablation_ratios}
    \small
    \setlength{\tabcolsep}{4pt}
    \begin{tabular}{lcccc}
        \toprule
        \textbf{Data } & \multicolumn{2}{c}{\textbf{Without CL}} & \multicolumn{2}{c}{\textbf{With CL}} \\
        \cmidrule(lr){2-3}\cmidrule(lr){4-5}
         & \textbf{MSE} & \textbf{RMSE} & \textbf{MSE} & \textbf{RMSE} \\
        \midrule
        Sparse regime I (80\%) & 0.6601 & 0.8125 & \textbf{0.5926} & \textbf{0.7698} \\
        Sparse regime II (60\%) & 0.6588 & 0.8117 & \textbf{0.6037} & \textbf{0.7770} \\
        Sparse regime III (50\%) & 0.8675 & 0.9314 & \textbf{0.7940} & \textbf{0.8911} \\
        \bottomrule
    \end{tabular}
\end{table}
In Sparse Regime I, which uses 80\% of the available training data, CL reduces MSE from 0.6601 to 0.5926 (10.2\%) and RMSE from 0.8125 to 0.7698. In Sparse Regime II, which uses 60\% of the available data, CL reduces MSE from 0.6588 to 0.6037 and RMSE from 0.8117 to 0.7770. In Sparse Regime III, which uses only 50\% of the available training data, CL reduces MSE from 0.8675 to 0.7940 (8.5\%) and RMSE from 0.9314 to 0.8911. Across all three regimes, CL consistently reduces both MSE and RMSE relative to the configuration without CL, indicating that its benefit holds regardless of the training data fraction used. As each regime uses a different test partition, absolute error magnitudes are not directly comparable across regimes. The consistent direction and relative magnitude of improvements within each regime are the primary evidence for CL's effectiveness under data scarcity.

\subsubsection{RMSE analysis}

\begin{figure}[t]
    \centering
    \includegraphics[width=0.30\textwidth]{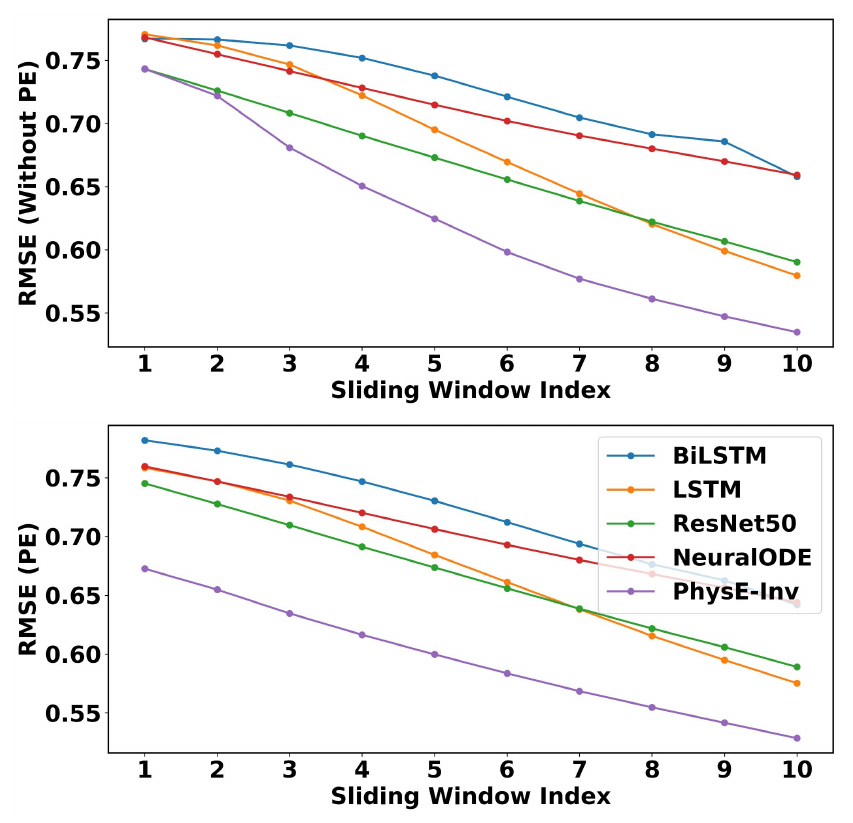}
    \caption{RMSE as a function of sequence position for 10-timestep sequence-to-sequence snow depth proxy estimation. Top panel shows models without parameter estimation (PE); bottom panel with PE.}
    \label{fig:rmse_sequence}
\end{figure}

Figure \ref{fig:rmse_sequence} shows the position-wise RMSE values for the sequence-to-sequence estimation task. Across all models, the error generally decreases toward the later positions in the sequence, indicating that additional temporal context helps improve estimation accuracy. PhysE-Inv achieves the lowest RMSE at every sequence position compared with all baseline models. The performance difference becomes more pronounced at later positions, suggesting that the model benefits from incorporating longer temporal dependencies. A comparison between the top and bottom panels further shows that the physics-encoded parameter estimation module consistently improves PhysE-Inv performance across the entire sequence, supporting the overall improvements observed in Table~\ref{tab:model_comparison}.

\subsubsection{Distributional analysis}

\begin{figure}[t]
\centering
\includegraphics[width=0.25\textwidth]{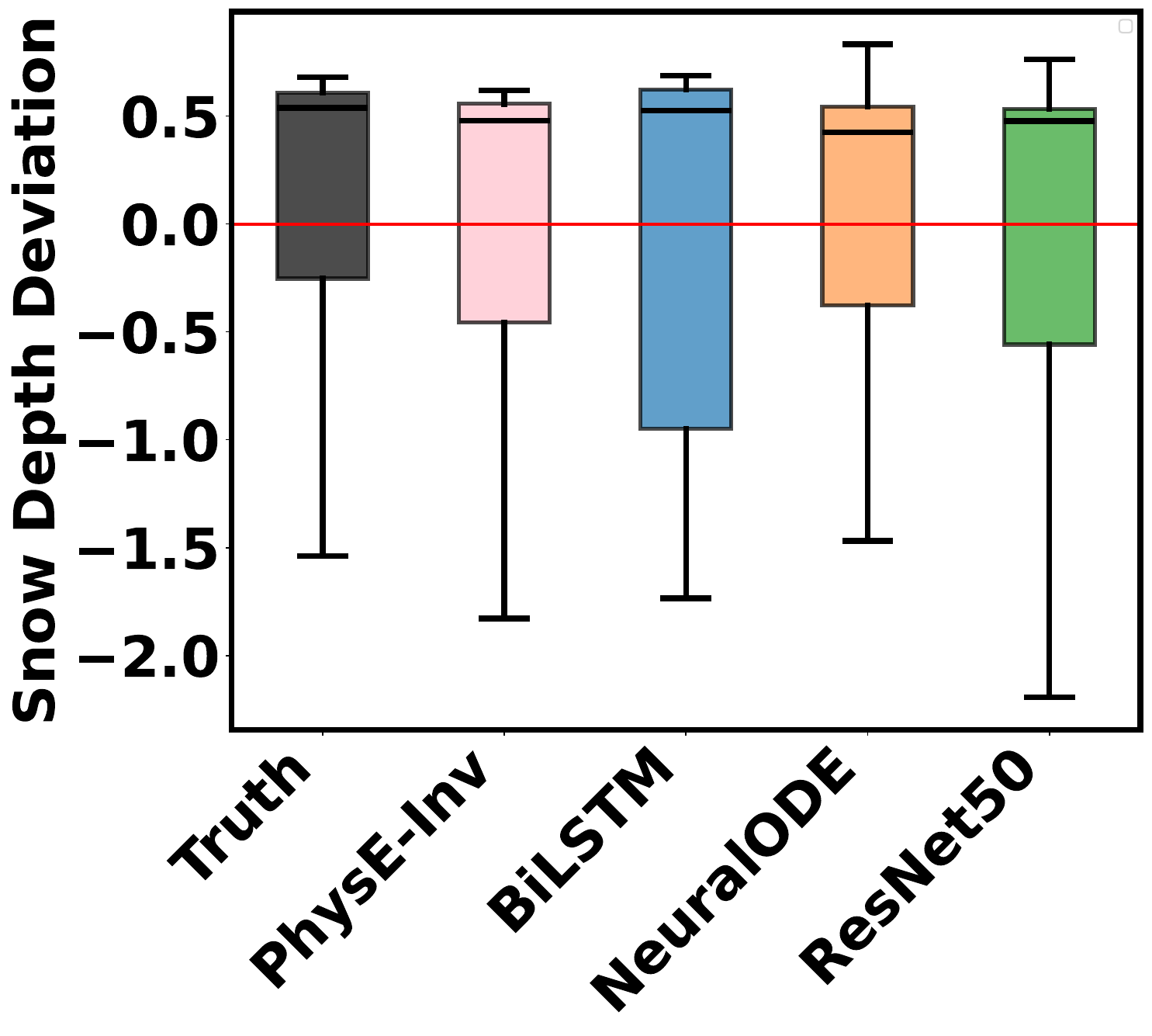}
\caption{Box-and-whisker plots comparing the distribution of estimated and proxy snow depth anomalies across different models. The red dashed line represents the zero-anomaly reference level.}
\label{fig:Boxplot_and_RMSE_wide}
\end{figure}

In addition to evaluating point-wise estimation errors, we examine whether PhysE-Inv can reproduce the statistical characteristics of the proxy snow depth anomalies. Figure \ref{fig:Boxplot_and_RMSE_wide} compares the distributions of estimated snow depth anomalies from different models using box-and-whisker plots across the test samples.

PhysE-Inv shows the closest agreement with the proxy target distribution among all evaluated models. Its median remains close to zero, and its interquartile range closely matches the spread of the proxy anomalies, indicating that the model captures both the central tendency and variability of the target distribution. In addition, PhysE-Inv produces fewer extreme outliers, suggesting more stable and consistent estimates.

Among the baseline models, BiLSTM achieves a median close to zero but exhibits a wider spread than the proxy target, particularly toward negative values, indicating greater variability in its estimates. ResNet50 produces a narrower distribution, suggesting that it does not fully capture the variability present in the proxy anomalies. NeuralODE exhibits a positive bias, with its median shifted above zero, and produces more extreme outliers, indicating less stable estimation behavior. Overall, the distributional analysis demonstrates that PhysE-Inv not only reduces point-wise estimation errors but also better preserves the statistical characteristics of the proxy snow depth anomalies.

\section{Conclusion and Future Work}

We present PhysE-Inv, a physics-encoded inverse modeling framework for estimating unobserved quantities from sparse time series observations. Our approach addresses a critical challenge in Arctic climate modeling, where direct snow depth estimates over sea ice are not available in commonly used reanalysis products. Instead, we estimate a physically motivated snow depth proxy informed by, but not directly derived from, the hydrostatic balance relationship between sea ice and snow. 

PhysE-Inv is motivated by the challenge of inferring unresolved physical states from aggregated reanalysis data, where multiple sub-grid configurations can produce similar grid-cell averages. Instead of explicitly defining a forward physical operator, the framework incorporates physical knowledge through a structured prior that guides the estimation process. Specifically, PhysE-Inv learns three differentiable physics-encoded parameters that link snow density information with the estimated proxy, providing physical guidance while maintaining flexibility in the inverse modeling framework.

Experimental results show that PhysE-Inv outperforms all evaluated baseline models, achieving the best overall performance under the parameter estimation setting with an MSE of 0.3568 and an RMSE of 0.5973 against the proposed proxy target. Furthermore, contrastive learning improves PhysE-Inv performance across varying sparse data regimes, achieving up to a 10.2\% reduction in MSE under limited data availability. The framework improves proxy anomaly distributional agreement while leveraging physics-encoded structure and contrastive learning for sparse inverse estimation.

A key limitation of this study is that evaluations are performed against a self-constructed proxy target rather than independent snow depth observations. Therefore, the reported MSE and RMSE quantify agreement with the proposed proxy target and should not be considered direct measures of real-world snow depth accuracy. Future work will focus on validation using in situ, airborne, or satellite-derived observations and incorporating Bayesian uncertainty quantification to characterize uncertainty in the estimated parameters.

\section*{Acknowledgments}

This work was supported by NSF grants OAC-2118285 (IHARP) and OAC-1942714.

\end{document}